\documentclass[10pt,twocolumn,letterpaper]{article}

\usepackage{wacv}
\usepackage{times}
\usepackage{epsfig}
\usepackage{graphicx}
\usepackage{amsmath}
\usepackage{amssymb}
\usepackage{multirow}
\usepackage{floatrow}
\usepackage[dvipsnames]{xcolor}
\usepackage{cuted}
\usepackage{capt-of}

\def\wacvPaperID{xxxx} % Enter the WACV Paper ID here

\wacvfinalcopy % *** Uncomment this line for the final submission

\ifwacvfinal
\def\assignedStartPage{1} % *** Enter the assigned starting page number (instead of 9876)
\fi

\ifwacvfinal
\usepackage[breaklinks=true,bookmarks=false]{hyperref}
\else
\usepackage[pagebackref=true,breaklinks=true,colorlinks,bookmarks=false]{hyperref}
\fi

\ifwacvfinal
\else
\fi

\begin{document}

%%%%%%%%% TITLE
\title{Deep Photo Scan: Semi-Supervised Learning for dealing with the real-world degradation in Smartphone Photo Scanning}

\author{Man M. Ho\\
Hosei University\\
Tokyo, Japan\\
{\tt\small man.hominh.6m@stu.hosei.ac.jp}

\and
Jinjia Zhou\\
Hosei University\\
Tokyo, Japan\\
{\tt\small jinjia.zhou.35@hosei.ac.jp}
}

\maketitle
\begin{strip}\centering
\includegraphics[width=\textwidth]{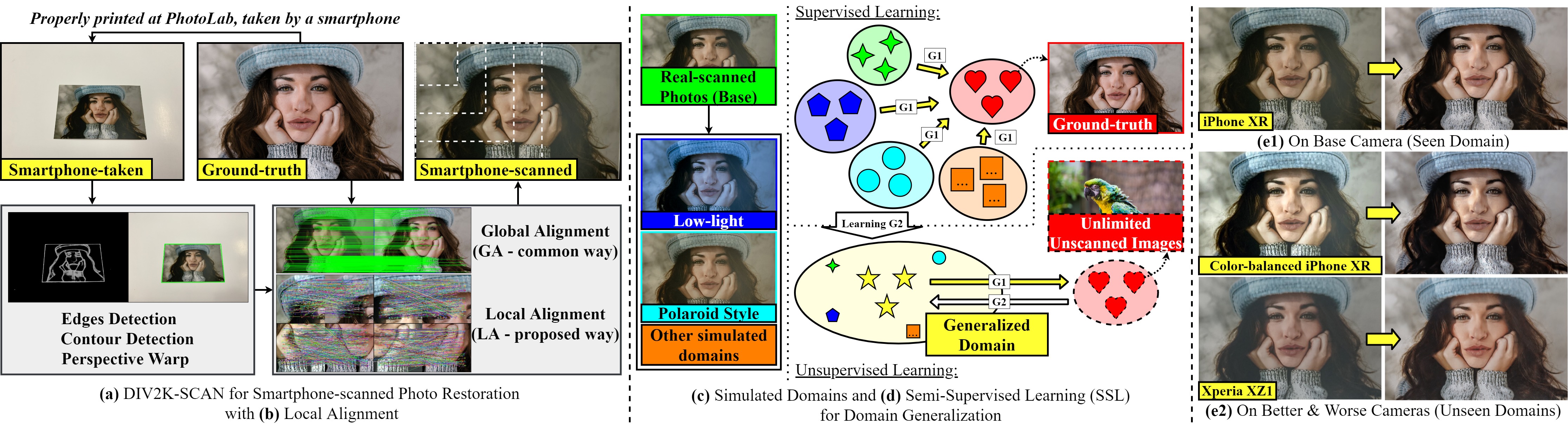}
\captionof{figure}{We present DIV2K-SCAN dataset for smartphone-scanned photo restoration (a) with Local Alignment (b), simulate varied domains to gain generalization in scanned image properties using low-level image transformation (c), and design a Semi-Supervised Learning system to train our network on also unscanned images, diversifying training image content (d). As a result, this work obtains state-of-the-art performance on smartphone-scanned photos in seen and unseen domains (e1-e2).
\label{fig:teaser}}
\end{strip}

%%%%%%%%% ABSTRACT
\begin{abstract}
Physical photographs now can be conveniently scanned by smartphones and stored forever as a digital version, yet the scanned photos are not restored well. One solution is to train a supervised deep neural network on many digital photos and the corresponding scanned photos. However, it requires a high labor cost, leading to limited training data. Previous works create training pairs by simulating degradation using image processing techniques. Their synthetic images are formed with perfectly scanned photos in latent space. Even so, the real-world degradation in smartphone photo scanning remains unsolved since it is more complicated due to lens defocus, lighting conditions, losing details via printing. Besides, locally structural misalignment still occurs in data due to distorted shapes captured in a 3-D world, reducing restoration performance and the reliability of the quantitative evaluation. To solve these problems, we propose a semi-supervised Deep Photo Scan (DPScan). First, we present a way of producing real-world degradation and provide the DIV2K-SCAN dataset for smartphone-scanned photo restoration. Also, Local Alignment is proposed to reduce the minor misalignment remaining in data. Second, we simulate many different variants of the real-world degradation using low-level image transformation to gain a generalization in smartphone-scanned image properties, then train a degradation network to generalize all styles of degradation and provide pseudo-scanned photos for unscanned images as if they were scanned by a smartphone. Finally, we propose a Semi-Supervised Learning that allows our restoration network to be trained on both scanned and unscanned images, diversifying training image content. As a result, the proposed DPScan quantitatively and qualitatively outperforms its baseline architecture, state-of-the-art academic research, and industrial products in smartphone photo scanning.
\end{abstract}

\begin{figure*}[ht]
    \centering
    \includegraphics[width=\textwidth]{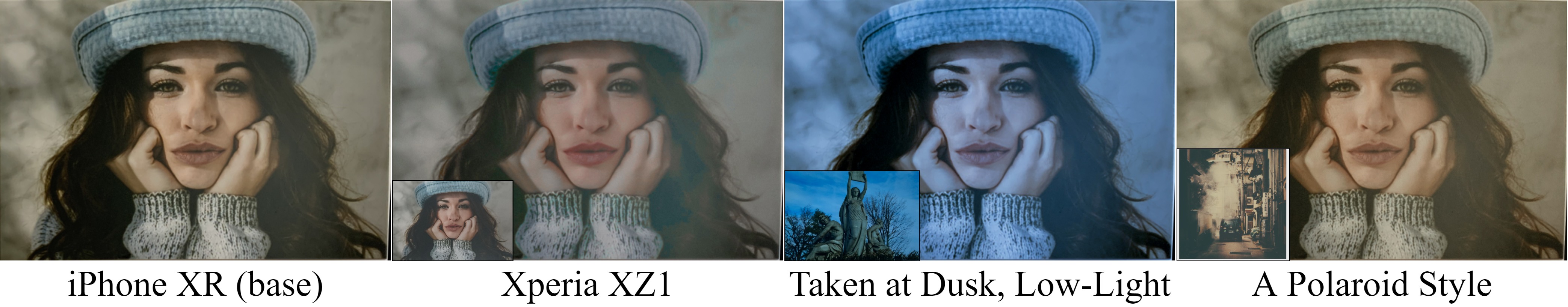}
    \caption{We manually simulate different domains of degradation affected by other shooting environments (\textit{taken at dusk with lack of light}) and devices (\textit{Xperia XZ1}, \textit{Polaroid Camera}) using low-level image transformation based on the real-world degradation. As a result, the simulated photos are qualitatively closed to the real-scanned photos, proving the feasibility of our approach in providing pseudo-scanned images for unscanned photos as if they were also taken in/by other shooting environments and devices.}
    \label{fig:poccsa}
\end{figure*}

%%%%%%%%% BODY TEXT
\section{Introduction}
Every moment passing by is precious; especially, when it marks important life events such as graduation, wedding. The moments are usually captured in photographs. However, we do not always have a digital version since digital cameras were not common in the past, or we accidentally lost it. Thanks to technological development, photographs can be efficiently stored by scanning applications on smartphones as high-resolution digital images. It also provides an efficient way to share the captured moments in physical photographs to everyone through the internet. To restore the scanned photos, the recent Old Photo Restoration (OPR) \cite{wan2020bringing, wan2020old} based on Deep Neural Networks (DNNs) tries to mimic and learn the scanning degradation using low-level image processing techniques such as Gaussian Noise, Gaussian Blur. Their artificially degraded photos are then formed with the perfectly scanned old photos in latent space. However, the smartphone-scanned photos are still not restored well since they contain more complicated degradation caused by camera quality, scanning environments, losing details via printing, various post-processing techniques, etc. In this work, we adopt DIV2K \cite{Timofte_2018_CVPR_Workshops} to present the DIV2K-SCAN dataset, which provides real-world degradation in smartphone photo scanning. Besides the common way of globally aligning a scanned photo to their ground-truth, we propose Local Alignment (LA) to reduce a minor misalignment remaining in data. Inspired by \cite{ho2021deep}, based on the captured real-world degradation, we simulate many different variants using low-level image transformation to gain a generalization in smartphone-scanned image properties for our restoration network. Furthermore, we leverage the concept of Generative Adversarial Networks (GANs) \cite{goodfellow2014generative, isola2017image, CycleGAN2017} to first generalize all domains of degradation, then provide pseudo inputs for an unlimited amount of unscanned images in training. Being joint with supervised training, we design a cycle process as high-quality images $\rightarrow$ scanned photos/pseudo inputs $\rightarrow$ reconstructed images. The proposed semi-supervised scheme balances two supervised and unsupervised errors while optimizing to limit the effect of imperfect pseudo inputs but still enhance restoration. Our approach is briefly described in Figure \ref{fig:teaser}. Besides, our code and data are available at \url{https://minhmanho.github.io/dpscan/}.

\textbf{Creating real-world degradation.}
The performance of DNNs mostly depends on how the training data is created. Therefore, it is crucial to create a specific degradation that is close to real-world problems. For example, \cite{Timofte_2018_CVPR_Workshops, wang2018esrgan, Dai_2019_CVPR, Li_2019_CVPR} use traditional interpolation techniques to achieve the distortion representing the problem of super-resolution. However, they are limited in digital zooming. \cite{zhang2019zoom} thus presented a way to obtain ground-truth for zoomed regions by optically zooming and provide a dataset for real-world computational zoom. Their work thus outperforms the state-of-the-art super-resolution work \cite{wang2018esrgan} in the field. \cite{plotz2017benchmarking,abdelhamed2018high} have the same motivation in solving the natural noise of photographs caused by low-ISO, \cite{ignatov2017dslr} enhancing smartphone-taken photos by creating training \{smartphone, DSLR camera\} pairs. This work is the first time to have a dataset (DIV2K-SCAN) that provides real-world degradation for smartphone-scanned photo restoration. Besides, inspired by \cite{ho2021deep}, we apply low-level image transformation to simulate varied variants of the captured degradation as if the photos were also captured in/by other shooting environments and smartphones for training. To demonstrate the feasibility of this scheme, we have provided a proof of concept shown in Figure \ref{fig:poccsa}. Our trained restoration network is thus generalized in smartphone-scanned image properties.

\begin{figure*}[t]
    \centering
    \includegraphics[width=\textwidth]{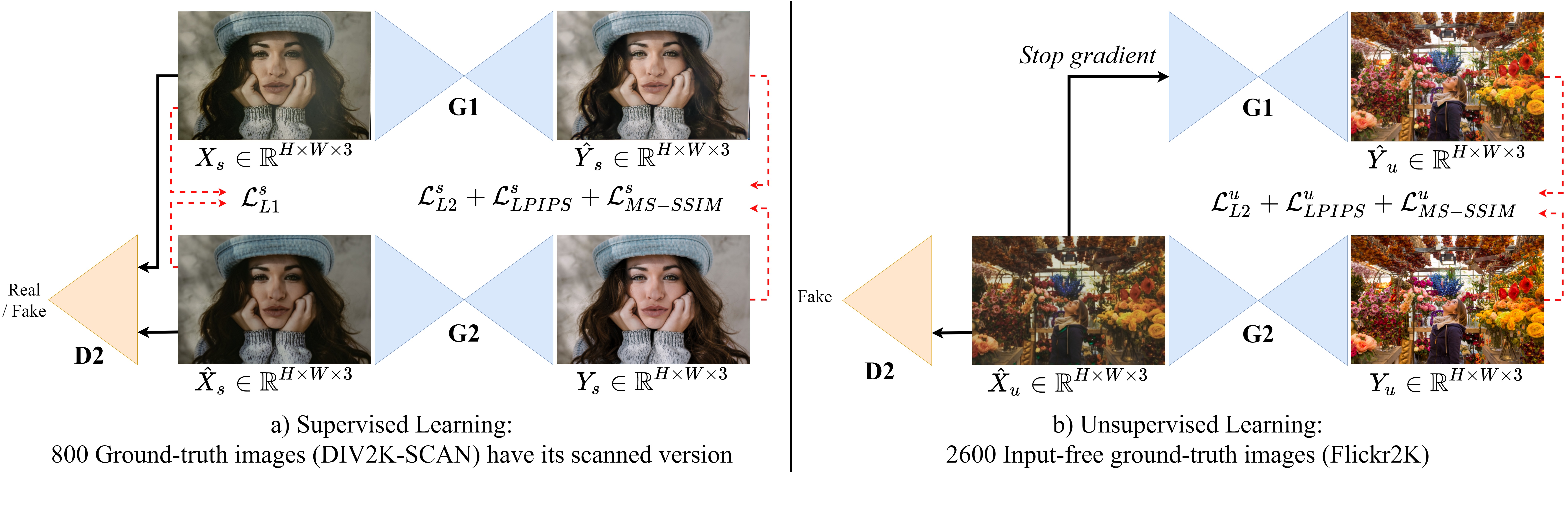}
    \caption{
    We present a Semi-Supervised Learning system that allows our model to be trained on scanned (supervised) (a) and unscanned (unsupervised) (b) photos under strongly supervised loss functions such as $L2$ ($\mathcal{L}^{*}_{L2}$), LPIPS \cite{zhang2018unreasonable} ($\mathcal{L}^{*}_{LPIPS}$), and MS-SSIM \cite{wang2003multiscale} ($\mathcal{L}^{*}_{MS-SSIM}$), where $*$ denotes $s$ or $u$ representing supervised or unsupervised scheme respectively. Meanwhile, the distribution of the real-scanned photos is captured with adversarial losses and $L1$ ($\mathcal{L}^{s}_{L1}$). Errors between (a) and (b) are balanced during optimization.
    }
    \label{fig:train_concept}
\end{figure*}

\textbf{Semi-Supervised Learning}.
Since the human-annotated data costs a considerable resource, worldwide researchers have proposed many semi-supervised learning techniques to solve the lack of labeled data (ground-truth) while training a deep neural network. For example, the image classification works \cite{semi_berthelot2019mixmatch,semi_zhai2019s4l,semi_chen2020big} utilized a well-trained model (teacher) to generate pseudo labels for unlabeled images so that the smaller network (student) can be trained on them. The semantic segmentation work \cite{ouali2020semi} added an unsupervised loss on synthetic maps predicted by auxiliary decoders for unlabeled samples. They all provided good schemes when the ground-truth data is limited. However, in this work, we lack the input data for training. Thanks to the seminal work of Generative Adversarial Networks (GANs) \cite{goodfellow2014generative}, the synthetic images now are high-fidelity and closer to the target distribution \cite{isola2017image}. Moreover, the synthetic images can be translated back to the input domain, creating a cycle consistency \cite{CycleGAN2017, wang2020semi} for the task. Inspired by the aforementioned works, we adopt the concept of GANs to generalize our real-world degradation and its simulated versions to map an unlimited amount of unscanned images to a generalized domain, providing them the pseudo-scanned inputs for unsupervised training. Thus, the training image content is diversified. Being joint with supervised training, it will create a cycle fashion as high-quality images $\rightarrow$ scanned photos/pseudo inputs $\rightarrow$ reconstructed images, as shown in Figures \ref{fig:teaser} and \ref{fig:train_concept}. The proposed semi-supervised scheme balances the errors between supervised and unsupervised while optimizing to limit the effect of imperfect pseudo inputs but still enhance restoration.

\textbf{Network architecture}. 
The autoencoder architecture U-Net \cite{ronneberger2015u} and its variants have gained high performance and become well-known in image-to-image translation tasks \cite{badrinarayanan2017segnet, zhang2017real, isola2017image, ho2019respecting}. Recently, many techniques for customizing a deep neural network have also grown rapidly towards enhancing efficiency and effectiveness. It is meaningful for this smartphone photo scanning to operate on a limited resource. 
Inspired by the attention mechanism, Wang et al. \cite{wang2020eca} proposed an Efficient Channel Attention (ECA) module that reduces a huge computational cost with favorable performance compared to its backbones in image classification, object detection, and semantic segmentation. Therefore, we leverage ECA to design a Residual ECA (RECA) Block and RECA U-Block for our restoration network. As a consequence, the customized architecture shows learning capability compared with its baseline architecture (named as Simple DPScan) built by U-Net \cite{ronneberger2015u}, residual modules \cite{johnson2016perceptual} between encoder and decoder, blur pooling \cite{zhang2019shiftinvar}, and EvoNorm-S0 \cite{liu2020evolving}. 

Our contributions are as follows:
\begin{itemize}
    \item We present the DIV2K-SCAN dataset, which provides real-world degradation in smartphone photo scanning. Also, the dataset allows a deep neural network to be trained with strongly supervised loss functions.
    \item Although the smartphone-scanned image pairs are aligned globally, a minor misalignment still occurs, lowering restoration performance and making a quantitative comparison using similarity metrics less reliable. To address this problem, we propose a Local Alignment (LA) to perfectly align a smartphone-scanned photo to its ground-truth. LA-ed data also shows that the larger the image size, the more serious misalignment.
    \item To address the concern of our performance on photos captured in/by other shooting environments and devices (generalization), inspired by \cite{ho2021deep}, we apply color style transfer to simulate varied types of degradation based on the real-world degradation. Thus, our work gains a generalization in smartphone-scanned image properties.
    \item We propose the semi-supervised Deep Photo Scan (DPScan) that has two advantages: a) Semi-Supervised Learning diversifying training image content by allowing our restoration network to be trained on both scanned and unscanned images and b) the customized Residual Efficient Channel Attention (RECA) Block and RECA U-Block. As a result, our semi-supervised DPScan outperforms its baseline, the previous research works, and industrial products comprehensively in 1-domain and generalization tests.
\end{itemize}

% \begin{itemize}
%  \setlength{\itemsep}{-1pt}
%     \item We present the DIV2K-SCAN dataset providing real-world degradation in smartphone photo scanning. A result shows that a simple supervised model trained on our dataset can outperform the previous work trained on the simulated degradation.
%     \item Even though the smartphone-scanned image pairs are aligned globally, the minor local misalignment still occurs, lowering restoration performance while training and making a quantitative comparison using similarity metrics less reliable. To address this problem, we propose a Local Alignment that smartphone-scanned photos are locally aligned.
%     \item We propose the semi-supervised Deep Photo Scan (DPScan) that has two advantages: 1) Semi-Supervised Learning enabling a deep model to be trained on both scanned and unscanned images and 2) the designed Residual Efficient Channel Attention (RECA) Block and RECA U-Block based on ECA \cite{wang2020eca}. As a result, our semi-supervised DPScan outperforms its baseline, the previous research work, and industrial products quantitatively and qualitatively in smartphone-scanned photo restoration.
% \end{itemize}

\section{The Proposed Deep Photo Scan (DPScan)}
Our work consists of two main components: (1) data preparation such as reproducing real-world degradation, image annotation/alignment, and simulating many smartphone scanning styles using color style transfer, and (2) the proposed semi-supervised DPScan for smartphone-scanned photo restoration.

Regarding (1), we leverage both traditional Canny \cite{canny1986computational} and DNN-based \cite{poma2020dense} edge detection techniques to identify the contour of interests. Afterward, the annotated images are warped and cropped to have a top-down view as though a professional scanner scanned them. Furthermore, we apply a precise alignment based on SIFT \cite{lowe2004distinctive} and RANSAC \cite{ransac_1, ransac_2} to suppress the structural mismatch between inputs and ground-truth images with Local Alignment (LA). Besides, inspired by \cite{ho2021deep} and the proof shown in Figure \ref{fig:poccsa}, we apply color style transfer to sample many different domains based on the real-world degradation as if the photos were also scanned in/by other shooting environments and devices, as shown in Figure \ref{fig:teaser}.
% from here
Regarding (2), we design a semi-supervised framework for our DPScan that includes two generators $G1$ and $G2$, and a discriminator $D2$. In that, $G1$ is to restore scanned photos and trained under supervised loss functions. Meanwhile, the GAN-based $G2$ provides the pseudo inputs for unscanned images in a generalized domain and is trained with the discriminator $D2$, which can distinguish whether a scanned photo is real or fake. Initially, all models are pre-trained on 1-domain DIV2K-SCAN (iPhone XR) with a supervised learning scheme, that $G1$ is trained independently with $G2$ and $D2$. Afterward, $G1$, $G2$, and $D2$ are jointly trained on scanned photos from DIV2K-SCAN and unscanned images from Flick2K \cite{Timofte_2018_CVPR_Workshops}, representing a Semi-Supervised Learning for dealing with the lack of inputs. In case of being fine-tuned on multiple-domain DIV2K-SCAN, $G2$ will generalize all domains and provide pseudo-scanned photos in the generalized domain, as shown in Figures \ref{fig:teaser} and \ref{fig:train_concept}. \textbf{Please check our supplemental document for a visualization of pseudo-scanned photos}.

\subsection{Supervised Learning for pre-training G1, G2 and D2}
\label{sec:supervised}
In pre-training on DIV2K-SCAN with supervised learning, after perspective warping, our $G1: X \rightarrow Y$ restores the scanned inputs $X_{s} \in \mathbb{R}^{H \times W \times 3}$ to have $\hat{Y}_{s} \in \mathbb{R}^{H \times W \times 3}$, as follows:

\begin{equation}
    \hat{Y}_{s} = G1(X_{s})
\end{equation}

The errors between $\hat{Y}_{s} \in \mathbb{R}^{H \times W \times 3}$ and its ground-truth images $Y$ are optimized under several supervised losses such as $L2$, Multiscale Structural Similarity (MS-SSIM) \cite{wang2003multiscale}, and the perceptual metric LPIPS \cite{zhang2018unreasonable} as follows:

\begin{equation}
\label{eq:g1_s}
    \mathcal{L}^{s}_{G1} = \alpha*\mathcal{L}^{s}_{L2} + \beta*\mathcal{L}^{s}_{LPIPS} + \gamma*\mathcal{L}^{s}_{MS\text{-}SSIM}
\end{equation}

where:

$\mathcal{L}^{s}_{L2} = ||Y_{s} - \hat{Y}_{s}||^2_2$,

$\mathcal{L}^{s}_{MS\text{-}SSIM} = MS\text{-}SSIM(Y_{s}, \hat{Y}_{s})$  described in \cite{wang2003multiscale},

$\mathcal{L}^{s}_{LPIPS} = LPIPS(Y_{s}, \hat{Y}_{s})$ described in \cite{zhang2018unreasonable}.

\begin{figure*}[t]
    \centering
    \includegraphics[width=\textwidth]{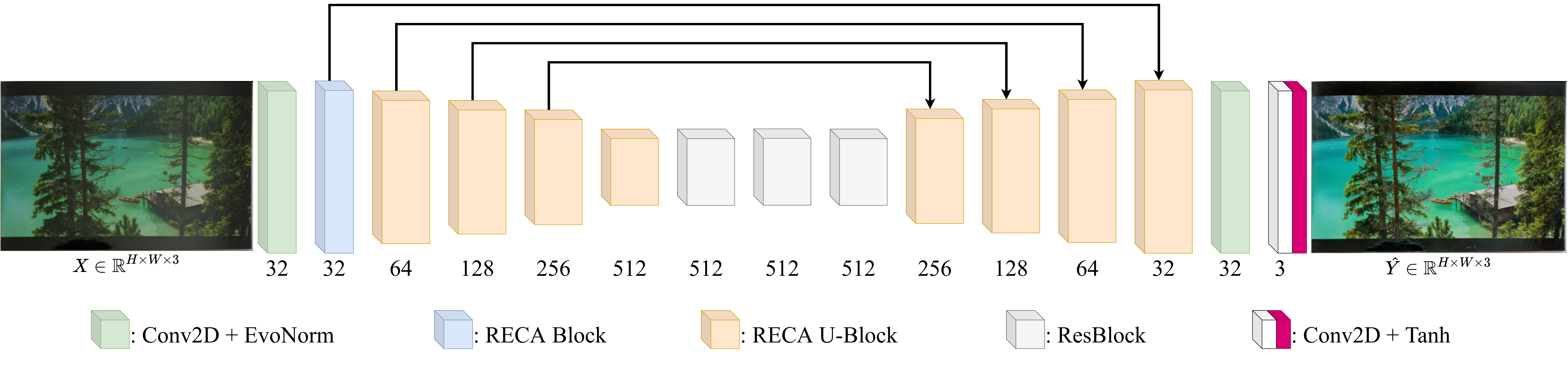}
    \caption{Architecture of our restoration network $G1$ restoring the scanned photo $X$ to have its high-quality $\hat{Y}$.}
    \label{fig:netg1}
\end{figure*}

\begin{figure}[t]
    \centering
    \includegraphics[width=\linewidth]{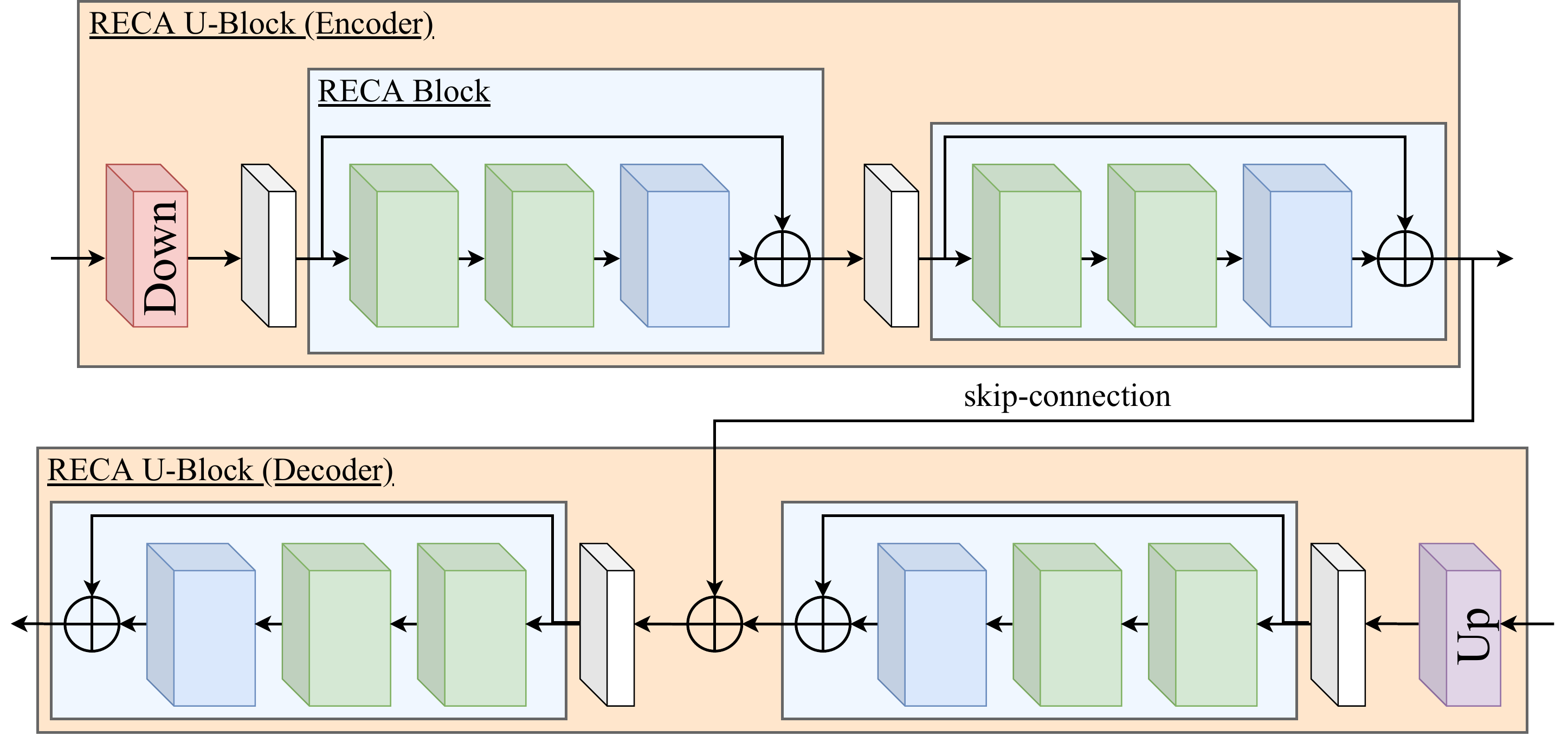}
    \caption{Illustrations of Residual Efficient Channel Attention (RECA) Block and RECA U-Block. Regarding "\textit{DOWN}"-/"\textit{UP}"-sampling, we use the anti-aliasing max pooling and bi-linear interpolation from \cite{zhang2019shiftinvar}. $\bigoplus$ represents a summation.}
    \label{fig:comg1}
\end{figure}

To learn the specific degradation of scanned photos from DIV2K-SCAN, we independently train a simply-designed network $G2: Y \rightarrow X$ to degrade the ground-truth images $Y_{s}$ to synthesize its scanned version $\hat{X}_{s} \in \mathbb{R}^{H \times W \times 3}$, as follows:

\begin{equation}
    \hat{X}_{s} = G2(Y_{s})
\end{equation}

% Since the utilized image alignment still consists of structural mismatch between $X_{s}$ and $Y_{s}$,
Instead of conditioning the discriminator \cite{miyato2018spectral, zhang2019self}, we train the generator $G2$ and discriminator $D2$ under the $L1$ \cite{isola2017image} and hinge adversarial losses \cite{lim2017geometric} as:

\begin{multline}
\label{eq:d2_s}
    \mathcal{L}^{s}_{D2} = - \mathbb{E}_{X_{s}}[min(0, -1 + D(X_{s})] \\
    - \mathbb{E}_{Y_{s}}[min(0, -1 - D(G2(Y_{s}))] 
\end{multline} 

\begin{equation}
\label{eq:g2_s}
    \mathcal{L}^{s}_{G2} = - \mathbb{E}_{Y_{s}}[D(G2(Y_{s}))] 
\end{equation}

\begin{equation}
\label{eq:L1_s}
    \mathcal{L}^{s}_{L1} = \mathbb{E}_{X_{s}, Y_{s}}[||X_{s}-G2(Y_{s})||_1]
\end{equation}

The total loss for G2 is defined as:

\begin{equation}
    \mathcal{L}^{s}_{G2\_final} = \alpha*\mathcal{L}^{s}_{L1} + \delta*\mathcal{L}^{s}_{G2}
\end{equation}

We empirically set $\alpha = 1$, $\beta = 0.2$, and $\gamma = 1$, $\delta = 0.05$ for this supervised learning scheme.

\subsection{Semi-Supervised Learning for fine-tuning G1, G2, and D2 together}

After pre-training models on DIV2K-SCAN, we then train $G1$, $G2$, and $D2$ together on both DIV2K-SCAN providing the ground-truth images $Y_{s}$ with their scanned photos $X_{s}$ and Flick2K \cite{Timofte_2018_CVPR_Workshops} providing the input-free ground-truth images $Y_{u} \in \mathbb{R}^{H \times W \times 3}$.

Firstly, $X_{s}$ and $Y_{s}$ are processed as described in Section \ref{sec:supervised}. Secondly, $G2$ provides pseudo inputs $\hat{X}_{u}$ for $Y_{u}$ as:

\begin{equation}
    \hat{X}_{u} = G2(Y_{u})
\end{equation}

under adversarial losses updated from Equations \ref{eq:d2_s} and \ref{eq:g2_s} as:

\begin{multline}
\label{loss_d2_su}
    \mathcal{L}_{D2} = - \mathbb{E}_{X_{s}}[min(0, -1 + D(X_{s})] \\
    - 0.5*( \mathbb{E}_{Y_{s}}[min(0, -1 - D(G2(Y_{s}))] \\
    + \mathbb{E}_{Y_{u}}[min(0, -1 - D(G2(Y_{u}))])
\end{multline}

\begin{multline}
\label{eq:g2_su}
    \mathcal{L}_{G2} = - 0.5*(\mathbb{E}_{Y_{s}}[D(G2(Y_{s}))]\\
    + \mathbb{E}_{Y_{u}}[D(G2(Y_{u}))] 
\end{multline}

Combined with the supervised loss in Equation \ref{eq:L1_s}, we have the updated total loss for $G2$ as:

\begin{equation}
    \mathcal{L}_{G2\_final} = \alpha*\mathcal{L}^{s}_{L1} + \delta*\mathcal{L}_{G2}
\end{equation}

Afterward, $G1$ leverages the pseudo input $\hat{X}_{u}$ to synthesize the reconstructed $\hat{Y}_{u}$ creating a cycle fashion as:

\begin{equation}
    \hat{Y}_{u} = G1(sg(\hat{X}_{u})) = G1(sg(G2(Y_{u})))
\end{equation}

where $sg$ denotes the $stop\_gradient$ function added to avoid falsifying the distribution of real-scanned photos. Similar to Equation \ref{eq:g1_s}, the defined loss function for $G1$ optimizing errors between $\hat{Y}_{u}$ and $Y_{u}$ is as: 

\begin{equation}
    \mathcal{L}^{u}_{G1} = \alpha*\mathcal{L}^{u}_{L2} + \beta*\mathcal{L}^{u}_{LPIPS} + \gamma*\mathcal{L}^{u}_{MS\text{-}SSIM}
\end{equation}

Finally, the semi-supervised loss function for $G1$ is as:

\begin{equation}
    \mathcal{L}_{G1} = \eta*\mathcal{L}^{s}_{G1} + (1-\eta)*\mathcal{L}^{u}_{G1}
\end{equation}

where $\eta$ is a balance weight between supervised and unsupervised errors, and is set to $0.5$. Other hyper-parameters empirically re-set $\alpha = 1$, $\beta = 0.1$, and $\gamma = 0.25$, $\delta = 0.05$ for this Semi-Supervised Learning scheme.

\subsection{Network architecture}
The proposed semi-supervised DPScan consists of three main deep neural networks: $G1: X \rightarrow Y$ for scanned photo restoration, $G2: Y \rightarrow X$ for degrading high-quality images, and discriminator $D2$ trained together with $G2$ to distinguish whether the scanned images are real or fake.

\textbf{Generator $G1$} is designed based on the network architecture of U-Net \cite{ronneberger2015u} with skip connections \cite{ho2019respecting}, residual modules (ResBlock) between the encoder and decoder \cite{he2016deep, johnson2016perceptual}, EvoNorm-S0 \cite{liu2020evolving}, which have achieved a high performance in image-to-image translation tasks. Besides, in each block of the encoder and decoder, we leverage the anti-aliasing max pooling and bi-linear interpolation \cite{zhang2019shiftinvar} for our down-sampling and up-sampling, respectively. The network architecture adopting the mentioned techniques is our baseline architecture, named Simple DPScan. Afterward, we adopt Efficient Channel Attention (ECA) module \cite{wang2020eca}, which has shown the efficiency but effectiveness in image classification, to design Residual ECA (RECA) and RECA U-Block and customize $G1$, as described in Figures \ref{fig:netg1} and \ref{fig:comg1}. Please find technical details in our supplemental document.

\textbf{Generator $G2$ and Discriminator $D2$}. We utilize the baseline architecture mentioned above for $G2$ to generate pseudo-scanned photos. Meanwhile, $D2$ is entirely based on the discriminator of SA-GAN \cite{zhang2019self} with Spectral Normalization \cite{miyato2018spectral}. Please find more details in our supplemental document.

\begin{figure*}[t]
    \centering
    \includegraphics[width=\textwidth]{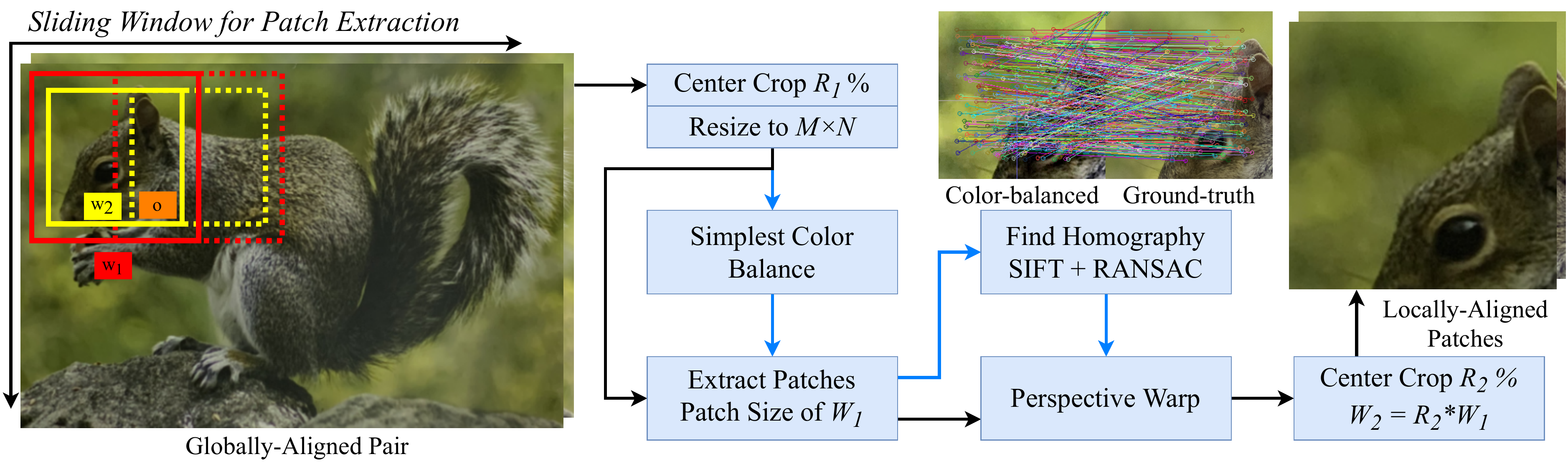}
    \caption{
    Generating Locally-Aligned data. After globally warping, we step-by-step apply a center crop to $R_1$\% of the current size to remove the black borders, resize to $M \times N$ using bicubic interpolation, extract patches from color-balanced \cite{limare2011simplest} photos to find homography matrices (\textcolor{blue}{blue}) and from original photos for warping using a sliding window with a size of $W_1$ and a stride of $S$\% of $W_1$, warp the scanned patches, center crop to $R_2\%$ of the size again, and finally obtain the locally-aligned patches with a size of $W_2=R_2*W_1$. $O=1-S/R_2$ denotes the percentage of how much two consecutive final patches overlap. Extracting and warping patches are powered by Kornia \cite{eriba2019kornia}.
    }
    \label{fig:patches}
\end{figure*}

\subsection{Data preparation and DIV2K-SCAN}
\label{sec:dp}
\textbf{Generating the real-world degradation}.
We first rotate the high-quality images from DIV2K \cite{Timofte_2018_CVPR_Workshops} so that all images are in landscape format, then center crop all images to an aspect ratio of $15:10$. Afterward, we ask a professional photography lab, where the staff is well-trained to print photographs with accurate colors and high quality, to print the processed images out with a size of $7.5 cm \times 5 cm$. All physical photos are then digitally taken in a room with sufficient light intensity on the white background using a smartphone. Therefore, our generated data contains a complex degradation of smartphone-scanned photos such as natural noises, haze, smartphone-level image quality, structural distortion, lost details via printing, etc.

\textbf{Contour detection and image alignment}.
After collecting the digital images of printed photos, we apply the efficient Canny \cite{canny1986computational} edge detection to detect the contours of the actual scanned photos in the white background. However, the method is sensitive to the color between the boundary and usually makes mistakes in detecting a complete contour. Thus, we utilize learning-based DexiNed \cite{poma2020dense} to support Canny's method. The remaining failed contour detection will be manually corrected by humans. \textbf{Please check our supplemental document for discussion and visualization}.

Even though the top-down view is obtained, the structural mismatch between the warped image and its ground-truth still occurs. It becomes more challenging to train a deep neural network. The recent work RANSAC-flow \cite{shen2020ransac} presents an advanced technique to precisely-structurally align an image to another one having the same context; however, the degradation of smartphone-scanned images can be falsified by their warping. To reduce structural mismatch while keeping the degradation intact, we leverage the SIFT features \cite{lowe2004distinctive} combined with RANSAC \cite{ransac_1, ransac_2} to find the homography, then align the warped images to their ground-truth. We eventually achieve $900$ ground-truth images with a real-scanned version for supervised training, validation, and test. Unfortunately, the local misalignment still occurs because the object shape of the photo taken in the 3-D world is usually distorted. Consequently, learning capability is harmed, and the quantitative evaluation using similarity metrics becomes less reliable. To address this issue, we present a way of generating Locally-Aligned (LA) data for training and test, as described in Figure \ref{fig:patches}.

\textbf{Training data}. We utilize $800$ scanned image pairs from DIV2K-SCAN and $2,600$ unscanned images from Flickr2K \cite{Lim_2017_CVPR_Workshops} to extract $12,000$ pairs and $20,800$ unscanned patches, respectively, using the presented Local Alignment (described in Figure \ref{fig:patches}) with $M \times N$ of $1080 \times 720$, $R_1$ of $95\%$, $R_2$ of $95\%$, final patch size $W_2$ of $256 \times 256$, the stride $S$ of $65\%$ $W_2$ resulting in two consecutive final patches overlapping $O=1-S/R_2 \approx 31.57\%$. Based on the real-world degradation, we synthesize more $K*12,000$ scanned photos as if they were also scanned in/by other environments and devices with a $K$ of $100$ color styles collected by \cite{ho2021deep}. In training, the data is augmented by a random flip in horizontal and vertical ways and random rotation with the degrees of $0, 90, 180, 270$.

\textbf{Validation and test data}. 
From $100$ globally-aligned images from a domain DIV2K-SCAN, we apply Local Alignment to extract $4000$, $1500$, $600$, $100$, $100$ patches with the final patch size $W_2$ of $176 \times 176$, $256 \times 256$, $384 \times 384$, $576 \times 576$, $1072 \times 720$, respectively, $M \times N$ of $1072 \times 720$, $R_1$ of $95\%$, $R_2$ of $80\%$, the stride $S$ of $50\%$ $W_2$ resulting in two consecutive final patches overlapping $O=1-S/R_2 \approx 37.5\%$. Regarding the final patch size $W_2$ of $1072 \times 720$, we only apply the first center crop and resize the photos. The pairs from each size is split into two sets, $40\%$ for validation (\textit{valset}) and $60\%$ for test (\textit{testset}). In each set, \textit{valset} contains degradation styles of iPhone XR and two simulated domains unseen from training, and \textit{testset} contains photos of iPhone XR, Color-Balanced \cite{limare2011simplest} iPhone XR, and Sony Xperia XZ1. In a comparison with industrial products, we set a $R_1$ of $85\%$ to avoid large black borders produced by Google Photo Scan.

\begin{figure}[t]
    \centering
    \includegraphics[width=\linewidth]{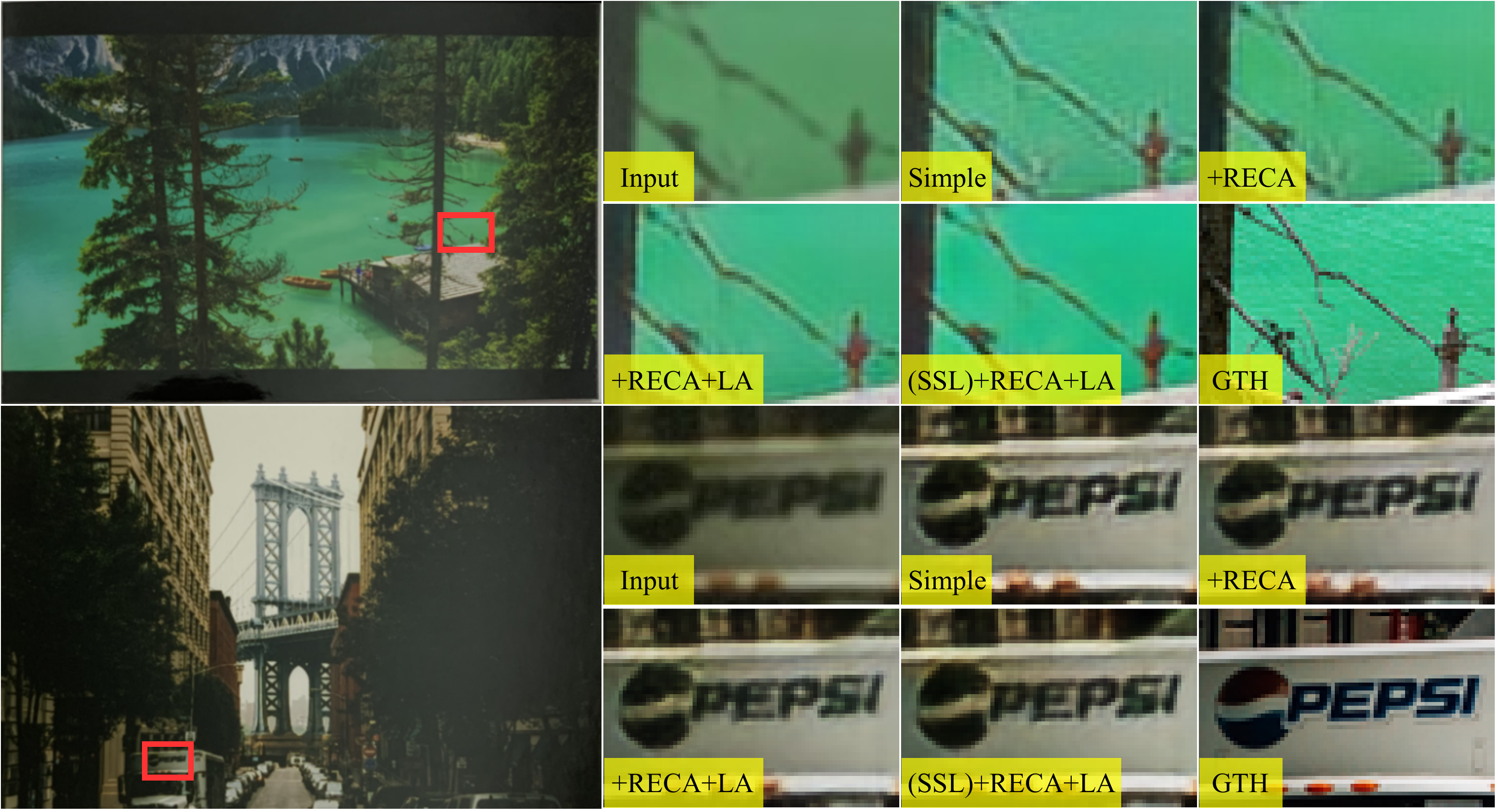}
    \caption{Ablation study on RECA for customizing DPScan, training on Locally-Aligned (LA) data, and the proposed Semi-Supervised Learning (SSL). As a result, each component has an improvement in restoring the edges and reducing artifacts. \textbf{Check our supplemental video for a better comparison}.}
    \label{fig:ab}
\end{figure}

\begin{table}[t]
\centering
\caption{Comparison between previous works \cite{isola2017image, CycleGAN2017} and our ablation models using (a) Supervised Learning (SL) or Semi-Supervised Learning (SSL), (b) Global Alignment (GA) or the proposed Local Alignment (LA), and (c) RECA. All models are trained and evaluated on 1-domain DIV2K-SCAN. $\uparrow/\downarrow$: higher/lower is better.}
\label{tab:ab_quan}
\resizebox{\linewidth}{!}{%
\begin{tabular}{|l|c|c|c|c|c|c|}
\hline
Method & Learning & Alignment & RECA & PSNR$\uparrow$ & LPIPS$\downarrow$ & MS-SSIM$\uparrow$ \\ \hline
Pix2Pix \cite{isola2017image} & SL & GA &  & 22.63 & 0.2138 & 0.8979 \\ \hline
CycleGAN \cite{CycleGAN2017} & SL & GA &  & 20.24 & 0.2504 & 0.8836 \\ \hline \hline
\multirow{5}{*}{1D-DPScan} & SL & GA &  & 23.78 & 0.1606 & 0.9275 \\ \cline{2-7} 
 & SL & GA & \checkmark & 24.10 & 0.1424 & 0.9316 \\ \cline{2-7} 
 & SSL & GA & \checkmark & 24.38 & 0.1423 & 0.9333 \\ \cline{2-7} 
 & SL & LA & \checkmark & 24.85 & 0.1351 & 0.9415 \\ \cline{2-7} 
 & SSL & LA & \checkmark & \textbf{25.26} & \textbf{0.1242} & \textbf{0.9446} \\ \hline
\end{tabular}%
}
\end{table}

\section{Experiments}
In this section, we conduct an ablation study on our presented techniques, including RECA for DPScan, training on Locally-Aligned (LA-ed) data, and the proposed Semi-Supervised Learning (SSL). Furthermore, we also train and compare with two typical works \textbf{Pix2Pix} \cite{isola2017image} and \textbf{CycleGAN} \cite{CycleGAN2017} in the same condition on 1-domain DIV2K-SCAN (iPhone XR). DPScan trained on only iPhone XR is denoted as 1D-DPScan.
Besides, to prove our generalization performance, we compare Generalized DPScan (G-DPScan) with the recent works that aim to solve many different degradation types of scanned photos such as \textbf {Industrial products Google Photo Scan (GPS) and Genius Scan (GS)} (we manually produce their results using iPhone XR), and \textbf{Academic research Old Photo Restoration} \cite{wan2020old} on also unseen sets of color-balanced \cite{limare2011simplest} iPhone XR and Xperia XZ1, which have better and worse performance than iPhone XR, respectively. Other experiments on deep modules for designing DPScan, image alignment, pseudo-scanned photo synthesis, the number of simulated domains $K$ can be found in the supplemental document.

\begin{figure*}[t]
    \centering
    \includegraphics[width=\textwidth]{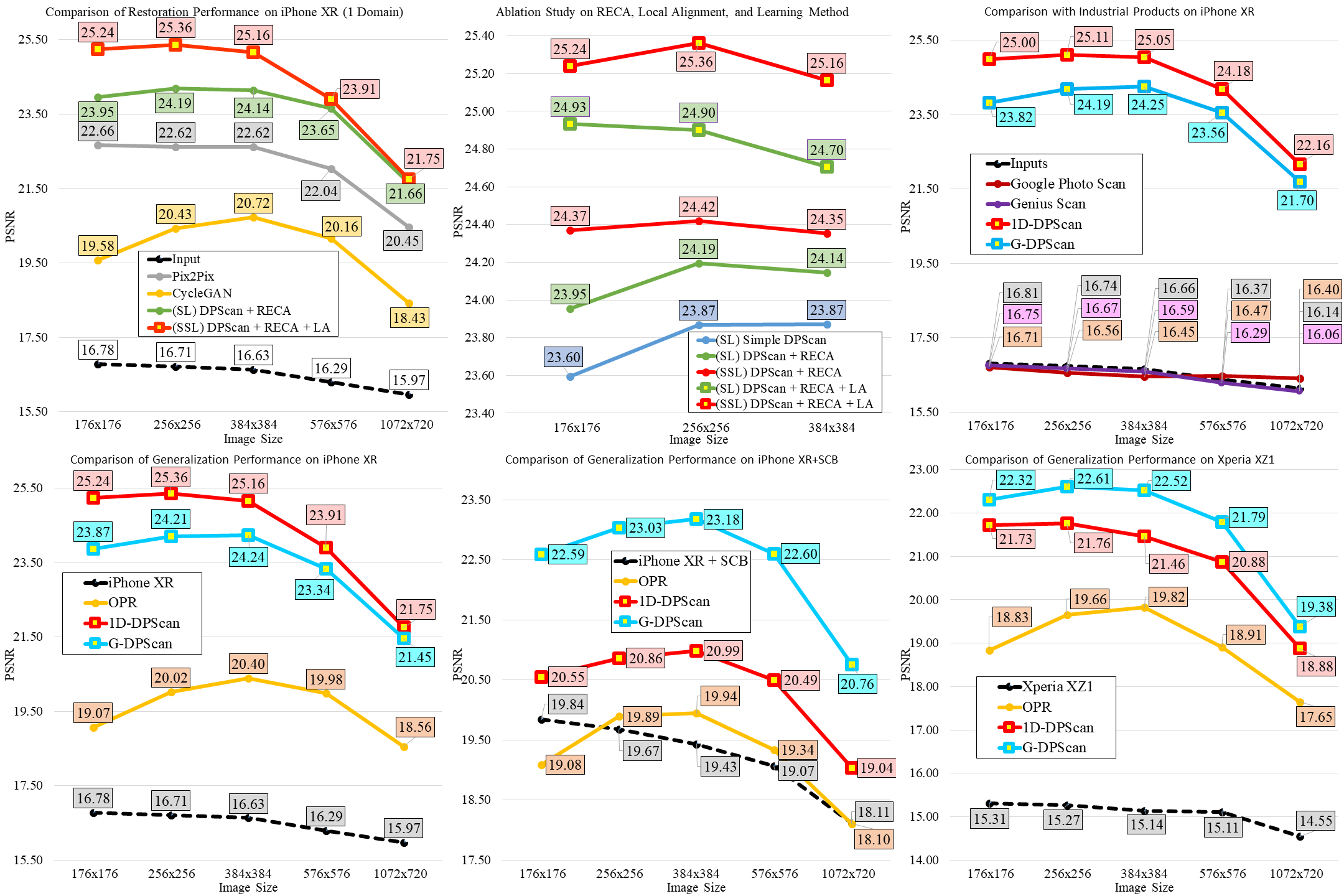}
    \caption{A full version of quantitative comparison using PSNR (\textit{higher is better}) on multiple-domain DIV2K-SCAN (iPhone XR is a seen domain, while iPhone XR + SCB \cite{limare2011simplest} and Xperia XZ1 are unseen domains) with an image size from $176 \times 176$ to $1072 \times 720$. Ablation models, 1D-DPScan, and the previous works Pix2Pix \cite{isola2017image} and CycleGAN \cite{CycleGAN2017} are trained on and to solve 1-domain DIV2K-SCAN (iPhone XR); meanwhile, other methods such as Old Photo Restoration (OPR) \cite{wan2020old}, industrial products, and our G-DPScan are to solve multiple domains. This experiment shows that 1) the image quality is gradually reduced in ascending order of image size, proving that the larger the image size, the more serious misalignment, 2) each presented technique provides a significant improvement, and the final version of DPScan outperforms all ablation models (\textit{middle-top}). 3) Our 1D-DPScan (trained on iPhone XR only) and G-DPScan (trained to solve multiple domains) outperform the research works \cite{isola2017image, CycleGAN2017, wan2020old} and industrial products Google Photo Scan and Genius Scan comprehensively. \textbf{Please check our supplemental document for a comparison using LPIPS and MS-SSIM}.}
    \label{fig:comp_quan_1md}
\end{figure*}

All quantitative comparisons are conducted on DIV2K-SCAN testsets described in Section \ref{sec:dp} using similarity metrics such as Peak Signal-to-Noise Ratio (PSNR), LPIPS \cite{zhang2018unreasonable}, and MS-SSIM \cite{wang2003multiscale}. Unfortunately, the similarity metrics are less reliable due to a minor misalignment remaining in data. Concretely, the smaller the image size, the smaller the structural mismatch between input and ground-truth after alignment, the more accurate the similar metrics, as shown in Figure \ref{fig:comp_quan_1md}. Therefore, we use the average score over three sizes of $176 \times 176$, $256 \times 256$, $384 \times 384$ for all quantitative comparisons on LA-ed data.

\begin{figure*}[ht]
    \centering
    \includegraphics[width=0.8\textwidth]{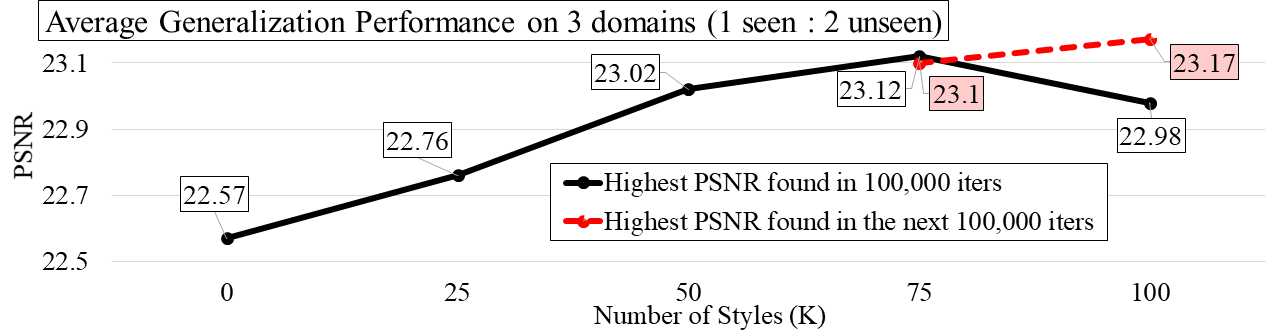}
    \caption{Ablation study on the number of simulated domains ($K$) for fine-tuning 1D-DPScan in two stages of $100,000$ iterations using average PSNR. The model with $K=100$ obtains the highest generalization performance, even though it takes a longer training time. \textbf{Please check our supplemental document for more details}.}
    \label{fig:nofstyles}
\end{figure*}

\begin{figure}[ht]
    \centering
    \includegraphics[width=\textwidth]{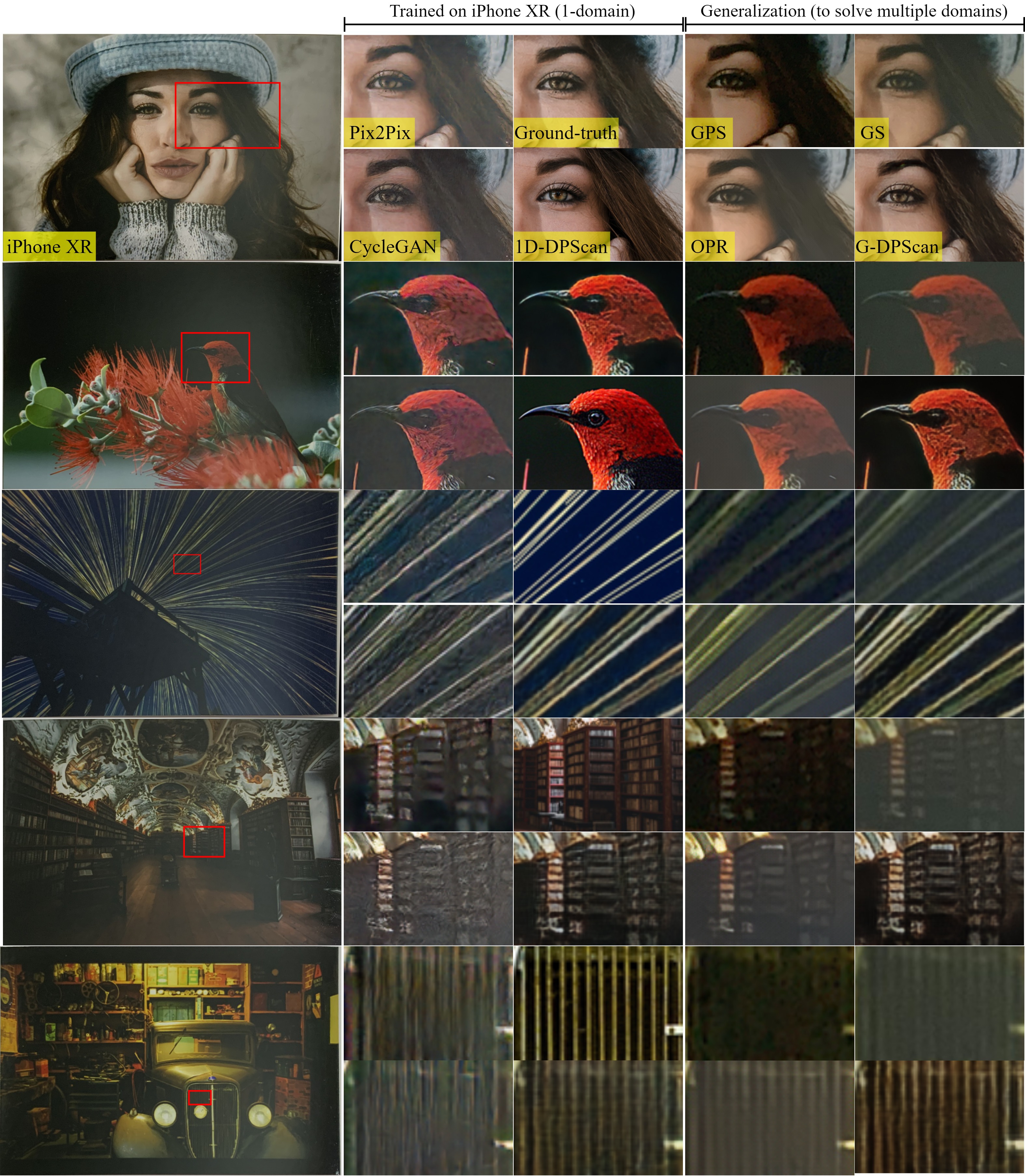}
    \caption{Qualitative comparison between two typical works Pix2Pix \cite{isola2017image} and CycleGAN \cite{CycleGAN2017} trained on 1-domain DIV2K-SCAN (iPhone XR), industrial products Google Photo Scan (GPS) and Genius Scan (GS), the previous work Old Photo Restoration (OPR) \cite{wan2020old}, and our 1-domain (1D-DPScan) and generalized (G-DPScan) networks. This work provides the most detailed photos without haze and color fading. \textbf{Please check our supplemental document for a full version and more results}.}
    \label{fig:comp_qua}
\end{figure}

\textbf{Comparison between ablation models and previous works trained and evaluated on 1-domain DIV2K-SCAN (iPhone XR)}.
\label{sec:ab}
We adopt U-Net \cite{ronneberger2015u}, residual modules \cite{he2016deep, johnson2016perceptual}, anti-aliasing down-/up-samplers \cite{zhang2019shiftinvar}, EvoNorm \cite{liu2020evolving} to design a baseline architecture, named Simple DPScan. While considering improving network architecture, we conduct an ablation study on customized deep learning techniques such as Flow Warping Block (FWB) \cite{ho2020sr}, Residual Feature-based Attention (RFA), Residual Self-Attention (RSA) \cite{zhang2019self}, Residual Channel Attention Block (RCAB) \cite{zhang2018image}, and Residual Efficient Channel Attention (RECA) \cite{wang2020eca}. The Efficient Channel Attention (ECA) \cite{wang2020eca} has shown the efficiency yet effectiveness in reducing computational costs with high accuracy for the image classification task. Moreover, an experimental result shows that the customized RECA outperforms other ablation techniques with the best image quality (technical and experimental details are in the supplemental document). Therefore, we leverage the RECA module and its variant RECA U-Block, which is customized for u-style architecture \cite{ronneberger2015u}, to design our network, as shown in Figures \ref{fig:netg1} and \ref{fig:comg1}. We also compare our ablation models with the Pix2Pix \cite{isola2017image} and CycleGAN \cite{CycleGAN2017} trained in the same condition to prove our restoration effectiveness. Besides, we propose Local Alignment and Semi-Supervised Learning (SSL) to solve (i) the remaining minor misalignment between the input and ground-truth in data and (ii) expensive costs leading to lack of real-scanned data. As a qualitative result, each presented component gradually improves the restoration performance as clearer edges with fewer artifacts, as shown in Figure \ref{fig:ab} (a better qualitative comparison is in our supplemental video). In comparison with previous works, our work provides the highest image quality without the haze effect, as shown in Figure \ref{fig:comp_qua}.
Quantitatively, our designed baseline architecture for DPScan can outperform the previous works \cite{isola2017image, CycleGAN2017} with better average PSNR, LPIPS \cite{zhang2018unreasonable}, and MS-SSIM \cite{wang2003multiscale} of \textbf{23.78, 0.1606, 0.9275}. The average \textbf{PSNR} is further improved \textbf{+0.32dB} when the baseline DPScan is customized with RECA, \textbf{+0.75dB} more when the model is trained on LA-ed data, \textbf{+0.41dB} more when the model is trained with the proposed SSL. Eventually, all presented techniques bring \textbf{+1.48dB} totally. Also, the average LPIPS and MS-SSIM are improved \textbf{-0.0364} and \textbf{+0.0171}, respectively, in total, as shown in Table \ref{tab:ab_quan} and Figure \ref{fig:comp_quan_1md}.

\begin{table*}[ht]
\centering
\caption{A quantitative comparison of generalization performance (1 seen: 2 unseen domains) between the research work OPR \cite{wan2020old}, our DPScan trained on iPhone XR only (1D-DPScan), Generalized DPScan (G-DPScan), and industrial products Google Photo Scan (GPS) and Genius Scan (GS). $\uparrow/\downarrow$: higher/lower is better.}
\label{tab:comp_quan}
\resizebox{\textwidth}{!}{%
\begin{tabular}{lccccccccccccllccc}
\cline{1-13} \cline{15-18}
\multicolumn{1}{|c|}{\multirow{2}{*}{Method}} & \multicolumn{3}{c|}{iPhone XR (seen)} & \multicolumn{3}{c|}{iPhone XR + SCB \cite{limare2011simplest} (unseen)} & \multicolumn{3}{c|}{Xperia XZ1 (unseen)} & \multicolumn{3}{c|}{Average} & \multicolumn{1}{l|}{} & \multicolumn{1}{c|}{\multirow{2}{*}{Method}} & \multicolumn{3}{c|}{iPhone XR (seen)} \\ \cline{2-13} \cline{16-18} 
\multicolumn{1}{|c|}{} & \multicolumn{1}{c|}{PSNR$\uparrow$} & \multicolumn{1}{c|}{LPIPS$\downarrow$} & \multicolumn{1}{c|}{MS-SSIM$\uparrow$} & \multicolumn{1}{c|}{PSNR$\uparrow$} & \multicolumn{1}{c|}{LPIPS$\downarrow$} & \multicolumn{1}{c|}{MS-SSIM$\uparrow$} & \multicolumn{1}{c|}{PSNR$\uparrow$} & \multicolumn{1}{c|}{LPIPS$\downarrow$} & \multicolumn{1}{c|}{MS-SSIM$\uparrow$} & \multicolumn{1}{c|}{PSNR$\uparrow$} & \multicolumn{1}{c|}{LPIPS$\downarrow$} & \multicolumn{1}{c|}{MS-SSIM$\uparrow$} & \multicolumn{1}{l|}{} & \multicolumn{1}{c|}{} & \multicolumn{1}{c|}{PSNR} & \multicolumn{1}{c|}{LPIPS} & \multicolumn{1}{c|}{MS-SSIM} \\ \cline{1-13} \cline{15-18} 
\multicolumn{1}{|l|}{Inputs} & \multicolumn{1}{c|}{16.71} & \multicolumn{1}{c|}{0.3734} & \multicolumn{1}{c|}{0.8436} & \multicolumn{1}{c|}{19.65} & \multicolumn{1}{c|}{0.3569} & \multicolumn{1}{c|}{0.8853} & \multicolumn{1}{c|}{15.24} & \multicolumn{1}{c|}{0.4327} & \multicolumn{1}{c|}{0.7783} & \multicolumn{1}{c|}{17.20} & \multicolumn{1}{c|}{0.3877} & \multicolumn{1}{c|}{0.8357} & \multicolumn{1}{l|}{} & \multicolumn{1}{l|}{Inputs} & \multicolumn{1}{c|}{16.74} & \multicolumn{1}{c|}{0.3944} & \multicolumn{1}{c|}{0.8367} \\ \cline{1-13} \cline{15-18} 
\multicolumn{1}{|l|}{OPR \cite{wan2020old}} & \multicolumn{1}{c|}{19.83} & \multicolumn{1}{c|}{0.3479} & \multicolumn{1}{c|}{0.8792} & \multicolumn{1}{c|}{19.64} & \multicolumn{1}{c|}{0.3498} & \multicolumn{1}{c|}{0.8786} & \multicolumn{1}{c|}{19.44} & \multicolumn{1}{c|}{0.4010} & \multicolumn{1}{c|}{0.8551} & \multicolumn{1}{c|}{19.64} & \multicolumn{1}{c|}{0.3662} & \multicolumn{1}{c|}{0.8710} & \multicolumn{1}{l|}{} & \multicolumn{1}{l|}{GPS} & \multicolumn{1}{c|}{16.58} & \multicolumn{1}{c|}{0.3916} & \multicolumn{1}{c|}{0.8431} \\ \cline{1-13} \cline{15-18} 
\multicolumn{1}{|l|}{1D-DPScan} & \multicolumn{1}{c|}{\textbf{25.26}} & \multicolumn{1}{c|}{\textbf{0.1242}} & \multicolumn{1}{c|}{\textbf{0.9446}} & \multicolumn{1}{c|}{20.80} & \multicolumn{1}{c|}{0.1883} & \multicolumn{1}{c|}{0.9172} & \multicolumn{1}{c|}{21.65} & \multicolumn{1}{c|}{0.2357} & \multicolumn{1}{c|}{0.8972} & \multicolumn{1}{c|}{22.57} & \multicolumn{1}{c|}{0.1827} & \multicolumn{1}{c|}{0.9197} & \multicolumn{1}{l|}{} & \multicolumn{1}{l|}{GS} & \multicolumn{1}{c|}{16.67} & \multicolumn{1}{c|}{0.3955} & \multicolumn{1}{c|}{0.8322} \\ \cline{1-13} \cline{15-18}
\multicolumn{1}{|l|}{G-DPScan} & \multicolumn{1}{c|}{24.10} & \multicolumn{1}{c|}{0.1413} & \multicolumn{1}{c|}{0.9363} & \multicolumn{1}{c|}{\textbf{22.93}} & \multicolumn{1}{c|}{\textbf{0.1610}} & \multicolumn{1}{c|}{\textbf{0.9276}} & \multicolumn{1}{c|}{\textbf{22.48}} & \multicolumn{1}{c|}{\textbf{0.2134}} & \multicolumn{1}{c|}{\textbf{0.9045}} & \multicolumn{1}{c|}{\textbf{23.17}} & \multicolumn{1}{c|}{\textbf{0.1719}} & \multicolumn{1}{c|}{\textbf{0.9228}} & \multicolumn{1}{l|}{} & \multicolumn{1}{l|}{G-DPScan} & \multicolumn{1}{c|}{\textbf{24.09}} & \multicolumn{1}{c|}{\textbf{0.1571}} & \multicolumn{1}{c|}{\textbf{0.9315}} \\ \cline{1-13} \cline{15-18} 
\multicolumn{13}{c}{a) w/ Recent Works. R1=95\% (R1: The first center crop ratio set to remove black borders.)} &  & \multicolumn{4}{c}{b) w/ Industrial Products. R1=85\%}
\end{tabular}%
}

\end{table*}

\begin{figure}[ht]
    \centering
    \includegraphics[width=\linewidth]{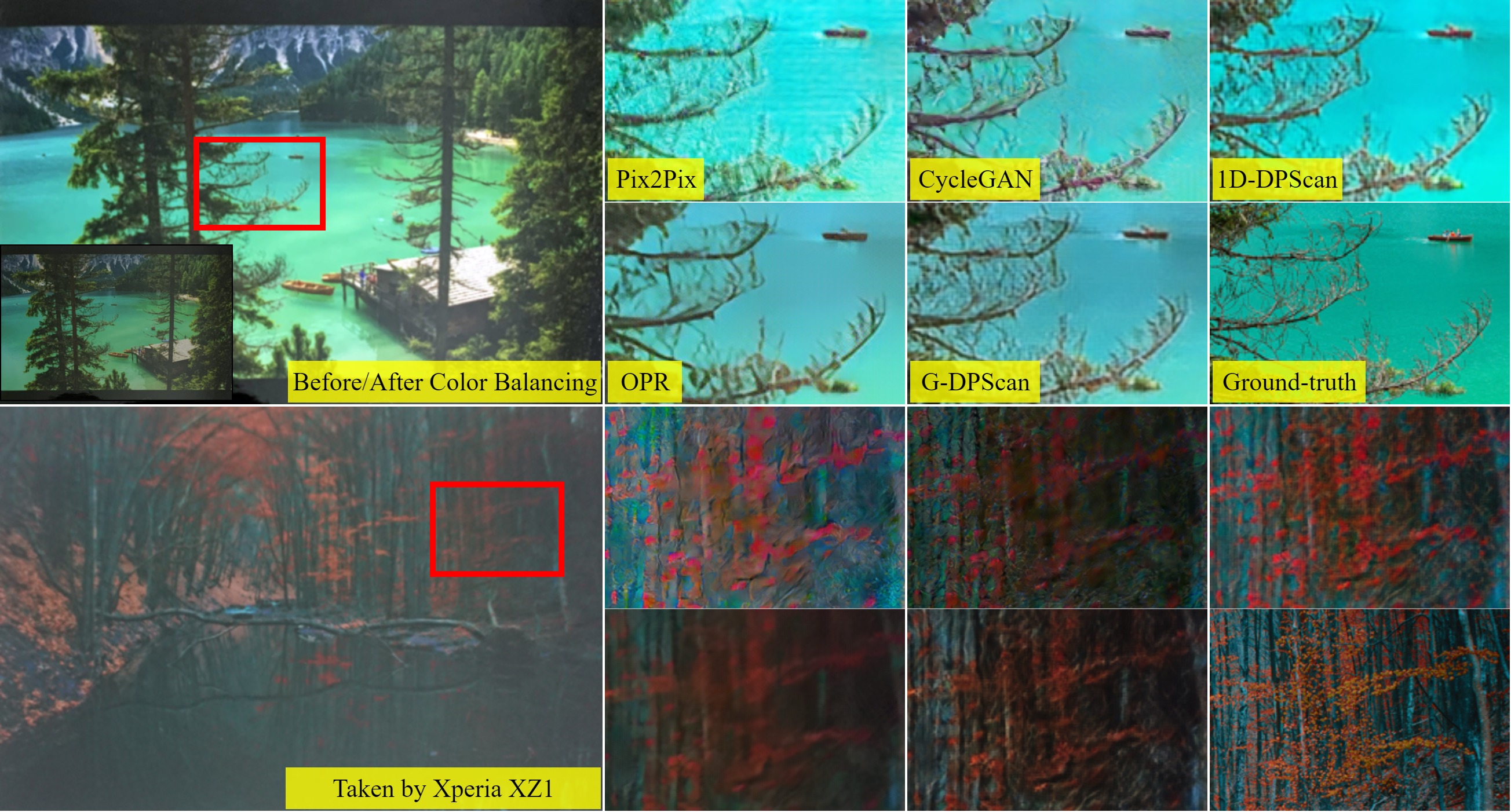}
    \caption{Qualitative comparison on unseen domains including the color-balanced \cite{limare2011simplest} iPhone XR (\textit{top sample}) and Xperia XZ1 (\textit{bottom sample}). Our 1-domain (DPScan) and generalized (G-DPScan) models obtains the highest image quality. \textbf{Please check our supplemental document for a full version and more results}.}
    \label{fig:ood_qua}
\end{figure}

\textbf{Comparison with previous works and industrial products on multiple-domain DIV2K-SCAN}. The recent works Google Photo Scan (GPS), Genius Scan (GS), and Old Photo Restoration (OPR) \cite{wan2020old} try to solve many different types of real-world degradation in smartphone photo scanning. Even though the industrial products GPS and GS have a user-friendly interface, their scanned photos are distorted with significant artifacts. Inspired by deep learning, OPR \cite{wan2020old} trained their network on the scanning degradation simulated by low-level image processing techniques. Although their synthetic images are formed with the perfectly real-scanned old photos in latent space, their model is weakly trained because the ground-truth images of real-scanned photos are missing. Moreover, the real-world degradation in smartphone photo scanning is more complicated caused by real lens defocus, lighting conditions, many different smartphone post-processing techniques, etc.

To overcome the aforementioned issues, we present DIV2K-SCAN providing real-world degradation, Local Alignment effectively reducing the remaining structural misalignment in data, RECA-customized architecture, scanned photo degradation simulation (inspired by \cite{ho2021deep}) for domain generalization in smartphone-scanned image properties, and Semi-Supervised Learning diversifying training image content by allowing our network to be trained on scanned and unscanned images.

As a quantitative result, our 1-domain DPScan (1D-DPScan) can outperform OPR with better average \textbf{PSNR, LPIPS, MS-SSIM} of \textbf{22.57, 0.1827, 0.9197}. However, 1D-DPScan is trained on iPhone XR only, and its performance on iPhone XR is much higher than on unseen photos from Simplest-Color-Balanced \cite{limare2011simplest} iPhone XR and Xperia XZ1, raising a concern of our generalization. Therefore, we simulate $K$ scanning domains based on our real-world degradation using color style transfer \cite{ho2021deep} as if our scanned photos were also taken in/by other environments and devices. An ablation study on $K \in \{25, 50, 75, 100\}$ shown in Figure \ref{fig:nofstyles} reveals that Generalized DPScan (G-DPScan) has the best generalization performance when $K=100$, even though it takes a long training time (more details are in the supplemental document). Eventually, our G-DPScan gains better average \textbf{PSNR, LPIPS, MS-SSIM} of \textbf{23.17, 0.1780, 0.9223} compared with 1D-DPScan and the research work OPR \cite{wan2020old}, and \textbf{24.24, 0.1615, 0.9330} compared with industrial products GPS and GS, as shown in Table \ref{tab:comp_quan} and Figure \ref{fig:comp_quan_1md}.

Qualitatively, our DPScan provides the clearest edges without haze effect in both 1-domain and generalization tests. Concretely, our results show the highest details of the \textit{girl}, \textit{bird}, and \textit{lines} on iPhone XR, as shown in Figure \ref{fig:comp_qua}. Regarding the domains unseen from training and validation, such as Simplest-Color-Balanced \cite{limare2011simplest} iPhone XR and Xperia XZ1, 1D-DPScan and G-DPScan provide the best image quality compared with the previous works. Surprisingly, 1D-DPScan generates more good-looking colors on its unseen domains than G-DPScan, although it is trained on iPhone XR only and quantitatively worse than G-DPScan, as shown in Figure \ref{fig:ood_qua}. \textbf{Please check our supplemental document for more interesting experiments and results}.
In conclusion, our semi-supervised DPScan outperforms the previous works and industrial products comprehensively.

\section{Conclusion}
We present a way to produce real-world degradation in smartphone photo scanning and present DIV2K-SCAN for smartphone-scanned photo restoration. Besides, Local Alignment is proposed to solve a minor misalignment remaining in data, which causes the reduction of restoration performance and reliability of the quantitative evaluation. Besides, we apply color style transfer to simulate many different variants of the real-world degradation as if the photos were also captured in/by other shooting environments and devices. Furthermore, we adopt the concept of GANs to degrade a high-quality image as if it were scanned by a smartphone and propose the semi-supervised Deep Photo Scan (DPScan) that has two advantages: 1) Semi-Supervised Learning allowing our network to be trained on both scanned and unscanned images and 2) the u-style architecture customized by Residual Efficient Channel Attention (RECA). Our work thus obtains a generalization in both training image content and smartphone-scanned image properties. As a result, our semi-supervised DPScan outperforms its baseline \cite{ronneberger2015u, zhang2019shiftinvar, liu2020evolving}, the industrial Google Photo Scan and Genius Scan, the recent work Old Photo Restoration \cite{wan2020bringing, wan2020old}, two retrained Pix2Pix \cite{isola2017image} and CycleGAN \cite{CycleGAN2017} in 1-domain and generalization tests quantitatively and qualitatively. This work, therefore, becomes a promising baseline for smartphone-scanned photo restoration.

{\small
\bibliographystyle{ieee_fullname}
\bibliography{egbib}
}

\end{document}